\documentclass[letterpaper,10pt,journal,twoside]{IEEEtran}
\usepackage{subcaption}
\usepackage{graphicx}
\usepackage{amsfonts}
\usepackage{amsmath}
\usepackage{booktabs}
\usepackage{xcolor}
\PassOptionsToPackage{hyphens}{url}
\usepackage{hyperref}
\title{Learning When a Robot Should Ask For Help:
Adaptive Intervention Requests in Human-Robot Teaming}

\author{
Yanran Jiang$^{1}$,
C\'ecile Paris$^{1}$,
Dana Kuli\'c$^{1,2}$,
and Leimin Tian$^{1}$%

\thanks{$^{1}$Yanran Jiang, C\'ecile Paris, Dana Kuli\'c, and Leimin Tian are with
CSIRO Technology, Australia.
{\tt\footnotesize yanran.jiang@csiro.au}}%

\thanks{$^{2}$Dana Kuli\'c is also with Monash University, Australia.}%

\thanks{\copyright~2026 IEEE. Personal use of this material is permitted.
Permission from IEEE must be obtained for all other uses, in any current
or future media, including reprinting/republishing this material for
advertising or promotional purposes, creating new collective works,
for resale or redistribution to servers or lists, or reuse of any
copyrighted component of this work in other works.}%

}

\begin{document}

\maketitle

\begin{abstract}

Robots operating in complex environments occasionally encounter situations where autonomous navigation fails. Requesting and receiving a timely intervention from their human supervisor can be critical for failure prevention or recovery. However, requesting help too frequently or at inappropriate times can increase operator workload and reduce team efficiency, while delayed requests may lead to prolonged downtime or unrecoverable failures. In addition to alert timing, the human, robot and task context also influences the intervention outcomes. This paper proposes an algorithm for learning to generate adaptive human intervention requests during human-robot teaming. A help-seeking policy is trained via offline reinforcement learning based on past intervention helpfulness and autonomous progress. This learned alert policy enables a robot to selectively request human help when the intervention outcome is likely to be beneficial. We evaluate the proposed policy in different task environments and user expertise levels in simulation, followed by evaluation in a physical user study. Our findings demonstrate that the learned policy improves exploration efficiency, reduces unnecessary alerts, and enhances novice user support without compromising expert performance in real-time human-robot teaming missions.

\end{abstract}

\section{Introduction}
Robots operating in complex, dynamic environments such as disaster zones, remote exploration sites, or industrial settings, often encounter situations that challenge their autonomy. For example, a robot navigating a tunnel may face unexpected debris, degraded sensing, or planner error that prevent continued progress \cite{kottege2024heterogeneous, petrlik2022uavs}. In such cases, human assistance can play an important role by providing corrective navigation guidance, such as suggesting alternative routes, placing waypoints to bypass obstacles, or helping the robot recover from localization or planning failures \cite{robinson2024human, chung2017ask, messina2005statement}.

Human operators with different levels of experience may vary both in their ability to decide when to intervene and in the quality of the interventions they provide \cite{conlon2022m,nanavati2021modeling, jiang2025influence}. Even expert operators may not always choose optimal timing or strategies for assisting a robot, while novice users may lack sufficient understanding of the robot’s internal state or capabilities to guide it effectively. Prior research has also identified occasions where human intervention was disadvantageous or unnecessary \cite{robinson2024human}. 

Users also differ in their preferences and responsiveness due to personal or contextual factors. Excessive or poorly timed requests can frustrate operators and disrupt their primary tasks or result in alert fatigue \cite{tariq2025alert}, while infrequent requests may lead to a loss in situational awareness \cite{senaratne2025framework}, inefficiency or failure to recover from difficult situations. Balancing these trade-offs is essential for effective human-robot collaboration \cite{nanavati2021modeling}.

How can a robot decide when to seek help? In many practical systems, alerts are triggered by static threshold-based rules (e.g., speed below a certain limit for a set duration) to detect problematic states. While simple and interpretable, these methods often fail to account for contextual factors such as environmental difficulty, the robot’s autonomous capabilities, or the effectiveness of past human interventions, which may vary across users and situations \cite{honig2018understanding}.

\begin{figure*}
    \centering
    \includegraphics[width=0.7\linewidth]{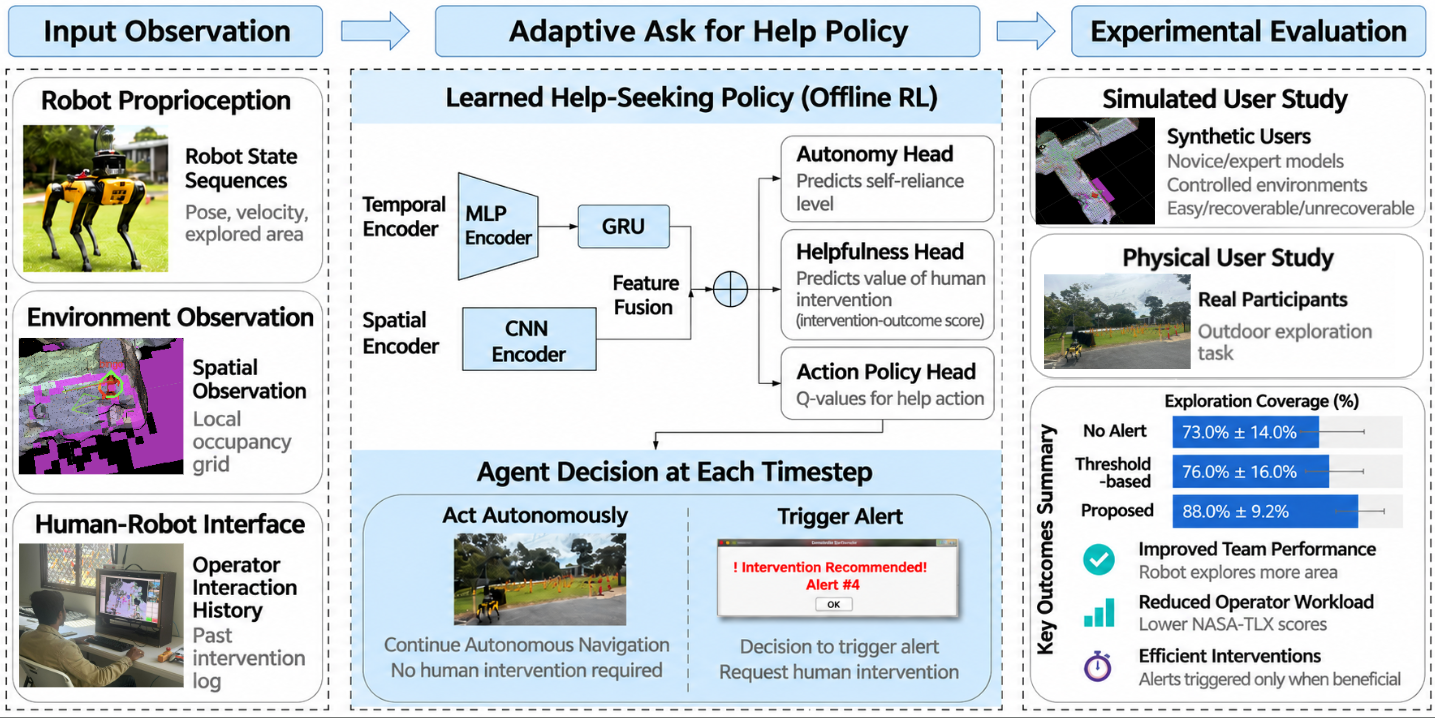}
    \caption{The proposed reinforcement learning framework for learning a help-seeking alert policy. Our model learns when to issue alerts based on environmental context, robot autonomy and human intervention utility, resulting in efficient alerts that improve task performance while reducing operator workload, as shown in simulated and physical user studies.}
    \label{fig:overview}

\end{figure*}

We propose an adaptive alert policy trained via offline reinforcement learning (RL), which learns to predict when a robot should request help based on historical data and current task context. Our method, illustrated in Fig.~\ref{fig:overview}, leverages two key signals: (1) intervention helpfulness, characterized by the exploration progress observed following an intervention, and (2) autonomy score, representing the robot's ability to make progress independently. By optimizing alerts for scenarios where human help is likely to be beneficial and the robot is unable to progress autonomously, our approach seeks to improve efficiency while reducing unnecessary interruptions—particularly important for novice users or high-tempo missions.

We conduct a simulated experiment and a physical user study comparing our learned alert policy against threshold-based, score-based, and LinUCB baselines across diverse environments and operator expertise levels\footnote{Implementation details, supplementary
results, and videos are available on the project page:
\url{https://human-robot-teaming.github.io/}}. Participants in the physical user study performed three exploration missions with different alert conditions, allowing us to assess the impact of alert strategy on exploration success, intervention dynamics, and team performance.  Our contributions are:

\begin{itemize}
    \item We propose a learning-based help-seeking policy that enables robots to decide when to request human assistance during exploration, trained via offline RL.
    \item We introduce a multi-head model architecture that jointly learns alert decisions and auxiliary predictions of autonomy and intervention helpfulness. These auxiliary supervised tasks provide additional learning signals that improve the policy’s ability to estimate recovery capability and intervention utility.
    
    \item We evaluate the proposed method in simulated and real-world experiments, demonstrating improved exploration performance, higher progress per alert, and reduced operator workload compared to a threshold-based baseline.

\end{itemize}

\section{Related Work}
This section reviews prior work across three interconnected themes that motivate our approach: how robots adapt behavior to different users in shared autonomy settings, how they handle the variability of human input, and how they decide when to request assistance.

\textbf{Shared Autonomy and User-Adaptive Robot Behavior:}
In shared autonomy, robots often operate alongside human partners, relying on timely assistance to resolve failures or improve task performance~\cite{yu2024encouraging,nanavati2021modeling}. 
Prior work has shown that adapting robot behavior to different users can improve collaboration and task performance, particularly in complex or uncertain environments \cite{nikolaidis2016formalizing}. A common approach is to model operator expertise  \cite{milliken2017modeling} or preferences based on interaction history or task performance~\cite{natarajan2024mixed}, and to adjust autonomy levels accordingly \cite{rabby2022learning}.

\textbf{Variability of Human Input:}
Human guidance may be delayed, inconsistent, or suboptimal due to limited situational awareness, cognitive load, or misunderstanding of robot capabilities~\cite{sridharan2011augmented,love2021should}. Prior approaches therefore estimate trust in feedback~\cite{love2021should}, weigh advice based on outcomes~\cite{loftin2016learning}, or learn whether to accept or reject human input~\cite{arakawa2018dqn,sridharan2011augmented}. However, these methods primarily assess input after it is provided rather than when it should be requested.

\textbf{When to Request Human Assistance:}
More recent work has explicitly examined when robots should request human assistance during task execution, recognizing that unnecessary or poorly timed requests can be costly. Chung et al.~\cite{chung2017ask} used robot introspection and policy expertise to trigger help requests when expected performance degrades. While this shows robots can autonomously identify when assistance may be needed, it assumes human input is uniformly beneficial and does not account for variation in intervention quality across task states or users. Other approaches focus on the robot's capabilities. Xie et al.~\cite{xie2022ask} requested assistance in potentially irreversible states, while Igbinedion and Karaman~\cite{igbinedion2024learning} used policy uncertainty to determine when help is needed. These approaches primarily base help-seeking on the robot's estimated competence, uncertainty, or risk. Contextual bandit approaches, such as LinUCB, have also been used to learn when to request assistance from the observed context~\cite{banerjee2024ask}. Complementary work models human factors such as helpfulness and busyness~\cite{nanavati2021modeling}, while operator behavior and intervention strategies have been shown to vary with expertise~\cite{jiang2025influence}.

\textbf{Summary and Gap:}
Overall, existing work suggests that effective human-robot collaboration depends not only on the availability of human input, but also on when and under what circumstances it is requested. However, these help-seeking approaches~\cite{chung2017ask,xie2022ask,igbinedion2024learning} do not explicitly account for how intervention effectiveness, which may be inconsistent or suboptimal~\cite{sridharan2011augmented,love2021should}, varies across task states or users. In contrast, our work focuses on learning help-seeking behavior that accounts for the state-dependent utility of human intervention as well as the robot's autonomous capabilities.

\section{Methodology}

In this work, we consider a robot exploring a challenging, partially unknown environment using semi-autonomous perception and navigation. Due to sensing uncertainty and environmental complexity, the robot may fail (e.g., get stuck or make inefficient progress). A remote human operator supervises the mission and can intervene by providing waypoint guidance, but continuous or poorly timed interventions increase workload without improving performance.

We aim to enable the robot to adaptively decide when to request human assistance, such that interventions are triggered only when likely to be beneficial. We formulate this as a decision-making problem and learn a policy that predicts the usefulness of intervention from the robot’s state and past interaction outcomes. The remainder of this section presents the system formulation and learning approach.

\subsection{Problem Formulation}
\label{sec:formulation}

We formulate robot help-seeking as a Partially Observable Markov Decision Process (POMDP) defined by the tuple
$(S,A,T,R,\gamma,\Omega,O)$. At timestep $t$, the environment is characterized by an underlying state $s_t\in S$, while the robot receives only a partial observation $o_t\in\Omega$ and selects an action
$a_t\in A=\{0,1\}$. Here, $a_t=0$ denotes continuing autonomous exploration and $a_t=1$ denotes requesting human assistance.

The underlying state captures three categories of information relevant to the help-seeking decision: (i) the robot's navigation condition, (ii) the local environmental condition, and (iii) latent factors determining the utility of requesting assistance in the current context. These latent factors include whether human assistance is likely to improve task progress and the human operator's expertise. The observation function $O(o_t\mid s_t)$ specifies the conditional distribution of observations given the underlying state, where $o_t$ contains robot motion, exploration progress, local costmap information, and observable interaction history.

To capture temporal dependencies, each observation is represented using a sliding window of the past \(K\) timesteps, i.e., $\mathcal{H}_t =
(o_{t-K+1},\ldots,o_t)$. This allows the policy to reason over trends in robot progress, such as stagnation or recovery. Rather than explicitly estimating the POMDP belief state, we learn an implicit belief representation:
$z_t=f_{\theta}(\mathcal{H}_t), z_t\in\mathbb{R}^{d}$, where $f_{\theta}$ combines recurrent encoding of the scalar observation history with convolutional encoding of the local costmap (Sec.~\ref{sec:implementation}).

To encourage $z_t$ to encode task-relevant outcome information, we introduce two auxiliary prediction objectives: an autonomous-progress head $\hat{A}_t = g_A(z_t)$ estimating the expected outcome under continued autonomous operation, and a helpfulness head $\hat{H}_t = g_H(z_t)$ estimating the expected outcome associated with human assistance. Both are supervised using retrospective targets from the logged trajectories, providing task-relevant supervision to the shared representation; the alert action itself is determined by the learned Q-values.

The transition function \(T(s_{t+1}\mid s_t, a_t, u_t)\) depends on both robot control and possible human intervention. If the robot triggers an alert (\(a_t\)=1),  the operator may provide an intervention \(u_t\) (e.g., a corrective waypoint), which influences the subsequent trajectory and next state \(s_{t+1}\). The reward function $R(s_t,a_t,u_t)$  rewards exploration progress, penalizes ineffective or unnecessary alerts, and rewards timely interventions that improve performance. We aim to learn a help-seeking policy \( \pi(a_t \mid z_t) \) that maximizes the expected return over a finite-horizon episode of length \( T \), using \(n\)-step temporal difference learning with discount factor \( \gamma \in [0,1] \):

\[
Q(z_t, a_t) \approx \sum_{i=0}^{n-1} \gamma^i r_{t+i} + \gamma^n \max_{a'} Q(z_{t+n}, a').
\]

Here, \( r_{t+i} \) is the observed scalar reward at step \( t+i \), \( \gamma \) discounts future rewards to reflect temporal preference, and \( n \) controls the bootstrapping horizon. The Q-value is bootstrapped from the predicted value of the state at time \( t+n \), using the maximum value over possible next actions \( a' \). To enable stable policy learning from offline data, we adopt a \textit{Conservative Q-Learning (CQL)} framework (see Sec.~\ref{sec:implementation}), which regularizes Q-values by penalizing overestimation on out-of-distribution actions not well represented in the dataset.

\subsection{Overview and Policy Architecture}
\label{sec:architecture}

We adopt an offline RL framework that trains a binary alert policy from logged robot behavior, intervention events, and mission outcomes. The policy network takes as input a sequence of scalar features (e.g., velocity, pose change, explored area, time since last alert) and a stack of local costmaps, enabling it to identify temporal patterns such as gradual stagnation or delayed recovery.

As shown in Fig.~\ref{fig:overview}, scalar inputs are encoded by an MLP followed by a GRU, while costmaps are encoded by a CNN. The resulting features are fused into a shared representation and passed to three parallel heads. Each head uses a fully connected hidden layer with ReLU activation. The Q-value head outputs two action values, while the autonomous-progress and helpfulness heads each produce a single sigmoid output.

\subsection{Auxiliary Predictive Models}
\label{sec:heads}

To provide additional context for alert decisions, we introduce two auxiliary heads: a human helpfulness predictor and a robot autonomy predictor. Both are trained via supervised learning on logged mission data. Their predictions serve as auxiliary learning targets that regularize the shared representation and help the policy reason about the expected outcomes of either requesting help or continuing autonomous operation.

\subsubsection{Human Helpfulness Predictor}
This module estimates the observed outcome associated with a human intervention given the current context (value range [0,1]). Rather than modeling operator willingness~\cite{nanavati2021modeling}, we ground prediction in the task-level outcomes: ground-truth labels are derived from post-intervention changes in explored area, where substantial coverage gains indicate a helpful intervention, and minimal progress indicates otherwise. The model learns to predict whether an intervention in the current context is likely to be followed by favorable exploration progress, based on the operator's intervention history and the robot's local state.

\subsubsection{Robot Autonomy Predictor}
This module estimates the robot's ability to make progress without assistance (value range [0,1]). Ground-truth scores are computed from autonomous operation intervals, excluding periods in which the robot executes user-provided waypoints after human intervention, to isolate its inherent exploration capability. The predictor learns to estimate this autonomous progress rate from local state and environmental observations, enabling the policy to anticipate whether continued autonomy is likely to be productive or not.

\subsection{Reward Function}
\label{sec:reward}

To guide the learning of the help-seeking policy, we design a reward function \( R(s_t, a_t, u_t) \) that reflects both per-step exploration performance and the impact of human interventions. It captures the following key components:

\begin{itemize}
    \item \textbf{Exploration progress:} At each timestep \( t \), the change in explored area is computed relative to the previous step. A reward of \( +1 \) is assigned if progress is observed, and \( -1 \) otherwise.
    
    \item \textbf{Alert-intervention outcomes:} When the robot triggers an alert (\( a_t = 1 \)) and a human intervention \( u_t \) follows, we compute the improvement in explored area before and after the intervention. If the intervention leads to measurable progress, a shaped reward is given:
    \[
    r_t = 1 + 10 \cdot \left(\frac{1}{1 + e^{-\Delta \text{area}}}\right),
    \]
    where \( \Delta \text{area} \) is the increase in the explored area post-intervention. If no progress occurs, a penalty of \( -1 \) is applied.
    
    \item \textbf{No intervention:} If an alert is triggered and no help is received, but the robot makes progress, a small penalty (\( -1 \)) is applied to reflect unnecessary alerting.  Conversely, if the robot remains stuck, a reward of \( +1 \) is assigned, indicating that the alert was appropriately triggered despite the absence of intervention.
\end{itemize}

This reward formulation encourages exploration progress while incentivizing necessary alerts that lead to effective human assistance.

\vspace{0.5em}

\subsection{Implementation and Training Details}
\label{sec:implementation}

The proposed policy is implemented as a deep Q-network with auxiliary prediction heads. 

We train the model offline using robot-human interaction data collected from 134 semi-autonomous navigation episodes in a high-fidelity simulated robot exploration environment. The dataset was originally introduced in~\cite{jiang2025influence} and contains intervention data collected during exploration tasks in unstructured tunnel-like environments. The system supports perception, mapping, and planning, enabling exploration in unstructured, tunnel-like environments. Human operators can intervene by providing waypoint guidance when the robot fails to make progress.

The dataset~\cite{jiang2025influence} includes four representative scenarios with annotated challenging regions, capturing common navigation difficulties such as narrow passages, intersections, dead ends, and uneven terrain. These scenarios induce a range of failure modes, including stagnation and obstacle-induced immobilization, requiring human intervention. To increase behavioral diversity, we further augment human waypoint-intervention data using the expertise-conditioned user model described in Section~\ref{sec:user_model}.

Each episode includes time-aligned state logs, costmaps, simulated human interventions, and mission outcomes. The final dataset contains over 99,000 training sequences, each consisting of 100 timesteps. These sequences are used to populate a replay buffer, from which fixed-length samples are drawn during training. Each sample includes the current and next state sequences, costmaps, actions, scalar rewards, and auxiliary supervision signals.

The training objective jointly optimizes the Q-values and auxiliary scores:

\[
\mathcal{L}_{\text{total}} = \mathcal{L}_{\text{Q}} + \lambda_{\text{aux}} \left(\mathcal{L}_{\text{help}} + \mathcal{L}_{\text{auto}}\right) + \lambda_{\text{CQL}} \cdot \mathcal{L}_{\text{CQL}},
\]

where \( \mathcal{L}_{\text{Q}} \) is a mean-squared Bellman loss using $n$-step temporal difference targets, \( \mathcal{L}_{\text{help}} \) and \( \mathcal{L}_{\text{auto}} \) are auxiliary prediction losses (\(L_2\) loss), and \( \mathcal{L}_{\text{CQL}} \) is a regularization term from CQL that penalizes overestimated Q-values for out-of-distribution actions. All losses are computed jointly, and no network component is pre-trained or frozen. In this work, $\lambda_{\mathrm{aux}} = $ 0.5, $\lambda_{\mathrm{CQL}} = $ 1.

We use the Adam optimizer with a learning rate of \( 10^{-4} \), a batch size of 64, and gradient norm clipping at 1.0. Target networks are updated every 10 epochs via hard synchronization. Scalar inputs are standardized using global statistics, and sigmoid normalization is applied to helpfulness and autonomy scores. The trained model was saved after offline training and deployed for inference on a local computer during evaluation.

\subsection{Baseline Policies}
\label{sec:baselines}

\subsubsection{Threshold-Based Alert Policy}

A rule-based system monitors robot motion and explored area. A stuck alert is
raised when displacement stays below 0.04\,m for more than 12\,s; an
inefficiency alert is raised when the explored area does not increase over a
10\,s window, indicating repeated traversal or slow progress despite continued motion. All thresholds were tuned by grid search on validation
episodes disjoint from the evaluation maps. Alerts use a generic message prompting operator intervention.

\subsubsection{Score-Based Alert Policy}
This baseline uses the auxiliary predictions without the RL Q-value head. It requests assistance when predicted autonomous progress is below $\tau_A$ and predicted intervention outcome is above $\tau_H$, with validation-selected thresholds $\tau_A=\tau_H=0.2$. Unlike the proposed policy, it does not optimize sequential return.

\subsubsection{LinUCB Contextual-Bandit Policy}
We adapt LinUCB \cite{banerjee2024ask} to the binary help-seeking decision using the same offline data, reward, and observable features as the proposed policy. The action maximising the LinUCB reward estimate and uncertainty bonus is selected. Validation gives $\alpha=0.5$ and ridge coefficient $\lambda=1.0$. Parameters remain fixed during evaluation, with no online updates.

\subsection{User Expertise Model}
\label{sec:user_model}

To simulate human interventions with varying expertise levels, we learn a data-driven model that predicts intervention waypoints from robot state and local environment context. The model takes as input the robot’s pose, velocity, local costmap features, and derived autonomy indicators, and outputs a continuous navigation target representing a human-provided intervention.

The model is trained on a user study dataset consisting of 27 participants, including both novice and expert operators~\cite{jiang2025influence}. Intermediate participants are retained and assigned to the novice or expert group according to their task performance in that study, and a separate regressor is trained for each group. The dataset captures human intervention behavior in robot exploration tasks, including waypoint corrections under different navigation conditions. To improve generalization given the limited dataset size, we apply data augmentation by perturbing intervention timing and robot state features.

Each regressor is implemented as a gradient-boosted multi-output regressor with 100 estimators, a learning rate of 0.05, a maximum tree depth of 3, and squared-error loss, achieving low prediction error (MAE = 0.063, MSE = 0.071). Because the prior study found that
robot-to-waypoint intervention distance differs with operator expertise~\cite{jiang2025influence}, we additionally construct an empirical distance distribution for each group. For each rollout, a user-specific
intervention distance is sampled from the corresponding distribution and held fixed for the rollout duration, so that rollouts within a group vary in intervention behavior while preserving the group-level differences observed in the real-user data.

\section{Evaluation}

We evaluate a learned alert policy for robot navigation using offline simulations (controlled benchmarking using simulated robot and user data) and an online physical user study (real-time evaluation with novice users in a human-robot team exploration task).

\subsection{Hypotheses}

\begin{itemize}
   
    \item H1: The learned alert policy improves team performance compared to the threshold-based alert policy.
 
    \item H2: The learned alert policy reduces operator workload during robot supervision tasks.

\end{itemize}

\subsection{Offline Simulation Evaluation with Simulated Users}\label{method_offline}

%\subsubsection{Experiment Design}

We first conducted an offline evaluation in a simulator to examine the behavior of the learned alert policy under controlled conditions. This simulation environment involves a human-robot team exploration task similar to the physical study (Section~\ref{method_online}) and the user expertise model training dataset (Section~\ref{sec:user_model}), in which the simulated user provides waypoint intervention when alerted by the robot. The simulation allows us to reduce variability in human response while preserving variation in intervention quality.

In this setting, simulated users respond immediately to the robot's alerts, removing variability in response timing. Differences in intervention quality are retained by modeling users with different expertise levels (novice and expert). Intervention waypoints are sampled from distributions of the learned user model (Section~\ref{sec:user_model}), allowing the simulation to represent both helpful and less helpful interventions. A total of 100 simulated interaction rollouts (50 novice-derived and 50 expert-derived rollouts) are generated to support comparisons across alert policies.

Evaluation is performed on a set of fixed maps that differ from those used during training or online physical user studies (details in supplementary material). Variations in map structure lead to different failure locations and intervention timings, resulting in diverse interaction trajectories.

Alerts are generated by the proposed policy, baseline policies described in Section~\ref{sec:baselines}, or ablated policy variants. The ablation removes the autonomy and/or helpfulness auxiliary objectives to assess their individual and combined contributions. These methods are evaluated under consistent conditions across three environment-difficulty levels and novice- and expert-derived intervention conditions. 
We additionally conducted a paired counterfactual analysis to assess whether post-intervention progress reflects a benefit of human assistance beyond autonomous recovery. We randomly selected 40 robot states and replayed each state with human intervention and autonomous continuation under closely matched initial conditions.

\subsection{Physical User Study}\label{method_online}

The study received approval from the institutional ethics committee. Twenty participants were recruited through invitations circulated within a research institution and a university, along with snowball sampling. Prior robot-supervision and navigation-based video-game experience was also recorded using four frequency levels (never to frequently). Participants first completed a training session supervising a Boston Dynamics Spot robot to explore an outdoor environment.  
The robot runs a field-tested subterranean-exploration autonomy stack~\cite{kottege2024heterogeneous}, comprising multi-agent 3D SLAM, multiple RGB and thermal cameras, a spinning lidar, a global frontier-based exploration planner, and a search-based local planner with obstacle avoidance and replanning. Prior to the user study, we conducted fully autonomous runs without human intervention to characterize where this stack degraded in the test environment. The physical environment was not designed to replicate the simulation geometry, but contained analogous robot-level challenges, including constrained passages, obstacles, navigation deadlocks, and exploration oscillations; representative physical failure cases are shown in the supplementary video.

Each participant completed three 10-minute exploration sessions under different alerting conditions: no alerts, threshold-based alerts, and adaptive alerts generated by the learned policy. The order of conditions was randomized across participants, with short breaks and brief questionnaires administered between sessions. Participants were instructed to supervise the robot naturally without prescribed strategies.

During each session, participants performed three concurrent tasks. They monitored the robot’s autonomous exploration using a map-based interface, responded to alerts when present by deciding whether or not an intervention was necessary, and placed waypoints to guide recovery if assistance was deemed necessary. The robot followed the participant's waypoint and then resumed autonomous exploration. In parallel, participants verified robot-detected objects by reviewing RGB images and associated labels, accepting or rejecting detections via keyboard input. Object verification occurred continuously throughout each session, introducing supervisory workload alongside navigation assistance.

\subsection{Measures}

We evaluate alert policies using both objective performance metrics and subjective user assessments. The following objective metrics are used:

\begin{itemize}
    \item \textbf{Total explored area}: The cumulative area covered by the robot during the mission, representing overall task performance.

    \item \textbf{Recovery time}: The duration between an alert and the robot’s resumption of effective exploration. In the simulation study, where user response timing is fixed, this metric primarily reflects intervention effectiveness. In the physical user study, it reflects both human response time and intervention effectiveness.

    \item \textbf{Progress per alert}: Total explored area divided by the number of alerts, measuring mission-level alert efficiency rather than the causal effect of individual alerts.

    \item \textbf{Post-alert progress}: The increase in explored area within a fixed horizon (T=10s) following an alert, capturing immediate post-alert progress while limiting unrelated later autonomous progress.
    \item \textbf{Alert success rate (physical user study only)}: The proportion of alerts that lead to a human intervention and measurable post-intervention exploration progress before the next alert or the end of the session.
\end{itemize}

In the physical user study, subjective measures were also collected to assess operator experience. After each trial, participants completed the NASA Task Load Index (NASA-TLX)~\cite{hart2006nasa} to measure perceived workload. Trust in the robot was assessed using Muir’s trust questionnaire~\cite{chen2018planning}. Participants also rated the perceived usefulness of alerts and completed a short qualitative questionnaire describing their preferred alert strategy and suggestions for improving alert timing or design.

\section{Results}

\subsection{Offline Simulation Results}

Results are summarized in Table~\ref{tab:offline_results} under expert- and novice-derived intervention conditions, aggregated across repeated trials. Statistical comparisons were conducted using Welch's two-sample $t$-tests, with 95\% confidence intervals and Cohen's $d$ effect sizes.

The proposed policy achieved the best overall performance across both conditions. Compared with the threshold-based, score-based, and LinUCB baselines, it achieved larger explored area (``Area''), shorter recovery time (``Rec. (s)''), and higher progress per alert (``Prog./Alert'')  and post-alert progress. Improvements in explored area and post-alert progress were significant against all baselines ($p<0.001$, $d=0.91$--$2.85$). The ablation results further show that the full model significantly improved explored area and post-alert progress over RL only under both conditions ($p<0.001$, $d=1.22$--$2.12$). The individual auxiliary objectives provided complementary and condition-dependent benefits, while their combination yielded the most consistent overall performance. The approximate paired counterfactual analysis further showed $7.7~\mathrm{m}^2$ greater exploration progress following intervention than autonomous continuation under closely matched conditions (95\% CI: $[6.5,9.0]$, $p<0.001$).

Overall, these results support H1 and demonstrate improved exploration performance and alert efficiency across intervention conditions.

\begin{table}[t]
\centering
\caption{Offline simulation performance and auxiliary-objective ablation
under expert- and novice-derived intervention conditions (mean $\pm$ s.d.).}
\label{tab:offline_results}
\setlength{\tabcolsep}{1.5pt}
\renewcommand{\arraystretch}{1.05}

\begin{tabular}{llcccc}
\toprule
\textbf{User} & \textbf{Policy} &
\textbf{Area} &
\textbf{Rec.} &
\textbf{Prog./Alert} &
\textbf{Post-Ale. Prog.} \\
& &
($\mathrm{m}^2$) &
(s) &
($\mathrm{m}^2$/alert) &
($\mathrm{m}^2$) \\
\midrule

Exp.
& Threshold
& $199.0{\pm}21.0$
& $70.5{\pm}19.3$
& $8.3{\pm}2.2$
& $7.1{\pm}2.4$ \\

& Score
& $218.6{\pm}12.7$
& $43.8{\pm}13.5$
& $14.6{\pm}2.4$
& $9.4{\pm}2.2$ \\

& LinUCB
& $224.3{\pm}10.9$
& $38.7{\pm}11.2$
& $16.8{\pm}2.1$
& $9.9{\pm}2.1$ \\

\cmidrule(lr){2-6}

& RL only
& $220.2{\pm}12.9$
& $40.4{\pm}11.8$
& $17.4{\pm}1.5$
& $10.5{\pm}2.1$ \\

& RL+A
& $224.5{\pm}10.7$
& $37.1{\pm}10.2$
& $18.1{\pm}2.1$
& $11.2{\pm}3.3$ \\

& RL+H
& $226.7{\pm}8.9$
& $35.2{\pm}9.1$
& $18.8{\pm}1.2$
& $12.0{\pm}5.1$ \\

& \textbf{Proposed }
& $\mathbf{232.0{\pm}4.5}$
& $\mathbf{29.1{\pm}10.4}$
& $\mathbf{21.1{\pm}2.5}$
& $\mathbf{15.0{\pm}4.1}$ \\

\midrule

Nov.
& Threshold
& $189.7{\pm}37.3$
& $100.6{\pm}27.9$
& $4.7{\pm}1.9$
& $5.3{\pm}2.2$ \\

& Score
& $205.4{\pm}14.8$
& $63.5{\pm}16.2$
& $10.8{\pm}1.7$
& $6.9{\pm}2.1$ \\

& LinUCB
& $205.8{\pm}13.1$
& $56.7{\pm}11.9$
& $13.2{\pm}1.5$
& $8.0{\pm}2.0$ \\

\cmidrule(lr){2-6}

& RL only
& $198.6{\pm}15.8$
& $45.7{\pm}10.4$
& $14.8{\pm}1.6$
& $6.7{\pm}2.5$ \\

& RL+A
& $215.9{\pm}13.6$
& $42.5{\pm}5.7$
& $15.5{\pm}0.8$
& $8.1{\pm}2.4$ \\

& RL+H
& $203.8{\pm}12.1$
& $43.9{\pm}6.8$
& $15.2{\pm}1.1$
& $9.5{\pm}3.3$ \\

& \textbf{Proposed}
& $\mathbf{217.0{\pm}10.4}$
& $\mathbf{41.2{\pm}3.2}$
& $\mathbf{16.7{\pm}0.1}$
& $\mathbf{12.0{\pm}2.5}$ \\

\bottomrule
\end{tabular}
\end{table}

\subsection{Physical User Study Results}
\begin{figure}
    \centering
    \includegraphics[width=\linewidth]{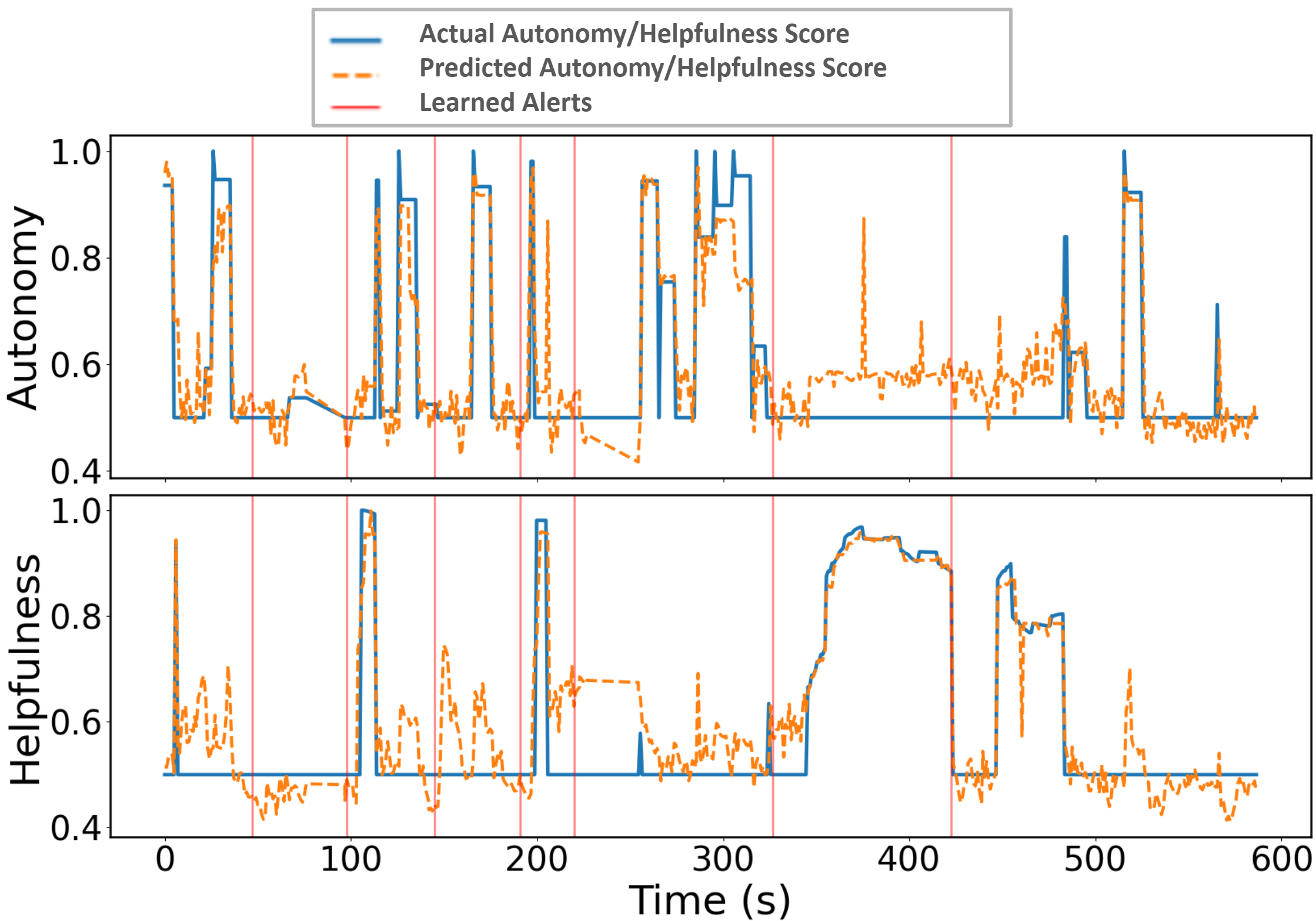}
    \caption{Predicted vs. Actual helpfulness and autonomy scores and the learned alert timings in an example user study trial.}
    \label{fig:trail}
\end{figure}

\subsubsection{Prediction Accuracy and Participant Adaptation}

Across all participants, the auxiliary prediction heads achieve low mean squared error (MSE) for helpfulness ($0.0091 \pm 0.0090$) and autonomy ($0.0072 \pm 0.0052$), indicating close alignment with target signals. Predicted helpfulness also matches the distribution of ground-truth values across participants (GT: $0.65 \pm 0.08$, Pred: $0.63 \pm 0.07$), suggesting that the model captures variability in intervention effectiveness across users.

Fig. \ref{fig:trail} shows the temporal behavior of the predictions in a representative trial. The model tracks the overall trends of both helpfulness and autonomy, with predicted signals closely following the ground truth. Alert events are generally aligned with transitions in these signals, suggesting that the learned policy responds to changes in robot autonomy and intervention effectiveness, despite the policy being trained in a simulated environment.

\begin{table*}[t]
\centering
\caption{Summary of physical user study results (mean $\pm$ SD). Statistical significance is evaluated using Friedman tests with post-hoc Wilcoxon signed-rank tests (Bonferroni-corrected). Effect sizes are reported using $r$ where available.}
\label{tab:online_results}
\begin{tabular}{lccc|c|c}
\toprule
\textbf{Metric} 
& \textbf{No Alert} 
& \textbf{Threshold} 
& \textbf{Proposed Adaptive Alert} 
& \textbf{$p$-value} 
& \textbf{Effect Size} \\
\midrule

Total explored area (m$^2$) 
& $569.38 \pm 108.63$ 
& $596.24 \pm 125.60$ 
& $\mathbf{678.33 \pm 121.90}$ 
& $p < .001$ 
& $W = 0.47$ \\

Number of alerts 
& -- 
& $17.0 \pm 6.30$ 
& $\mathbf{10.70 \pm 3.90}$ 
& $p= .042$ 
& $r = 0.43$ \\

Progress per alert (m$^2$/alert) 
& -- 
& $49.81 \pm 21.81$ 
& $\mathbf{73.12 \pm 35.19}$ 
& $p = .016$ 
& $r = 0.54$ \\

Recovery time (s) 
& -- 
& $14.01 \pm 11.90$ 
& $11.02 \pm 10.40$ 
& $p = .056$ 
& $r = 0.43$ \\

Alert success rate 
& -- 
& $0.47 \pm 0.23$ 
& $\mathbf{0.63 \pm 0.19}$ 
& $p = .048$ 
& $r = 0.43$ \\

NASA-TLX (workload) 
& $51.54 \pm 15.61$ 
& $43.78 \pm 14.21$ 
& $\mathbf{39.45 \pm 12.20}$ 
& $p < .001$ 
& $W = 0.53$ \\

Trust score 
& $3.35 \pm 0.89$ 
& $3.73 \pm 0.98$ 
& $3.48 \pm 0.91$ 
& $p = .331$ 
& $W = 0.06$ \\

Alert usefulness 
& -- 
& $5.21 \pm 0.55$ 
& $5.29\pm 0.59$ 
& $p = .850$ 
& $r = 0.23$ \\

\bottomrule
\end{tabular}
\end{table*}

\subsubsection{Objective Measures}

Table~\ref{tab:online_results} summarizes the user study results. The proposed adaptive alert policy achieved the highest explored area, significantly outperforming both threshold {($p_{\mathrm{adj}} < .001$)} and no-alert {($p_{\mathrm{adj}} < .001$)} baselines, supporting H1. It also generated fewer alerts than the threshold condition, indicating reduced operator interruptions while improving performance.
Regarding alert efficiency, the learned policy achieved higher progress per alert and alert success rate, suggesting more effective intervention timing. Recovery time showed a descriptive improvement but was not statistically significant.

\subsubsection{Subjective Measures}
As {summarized} in Table~\ref{tab:online_results}, perceived workload differs significantly across conditions, {with the learned alert policy yielding the lowest NASA-TLX scores, significantly lower than both the threshold ($p_{\mathrm{adj}} = .017$) and no-alert ($p_{\mathrm{adj}} = .001$)} conditions, supporting H2. In contrast, trust and perceived alert usefulness do not show significant differences across conditions.

Qualitative responses indicated that participants experienced difficulty dividing attention between robot supervision and the concurrent object classification task. Monitoring multiple interface panels increased cognitive load, and participants reported that alerts reduced the need for continuous manual monitoring, particularly during periods of high task demand.
Participants primarily relied on observable robot behaviours, such as becoming stationary, slowing down, or revisiting explored areas, as primary cues for intervention. Participants noted that alerts were most beneficial when their attention was occupied by the secondary task, as alerts acted as reminders to re-engage with the robot (e.g., ``Great to have assistance while I was verifying objects'' [P001]). These observations align with the quantitative results, where the learned alert policy reduced workload and the number of alerts while improving overall task performance.

\subsubsection{Understanding When the Robot Requests Help}

\begin{figure}[t]
    \centering
    \begin{subfigure}[t]{0.495\columnwidth}
        \centering
        \includegraphics[
            width=\linewidth
        ]{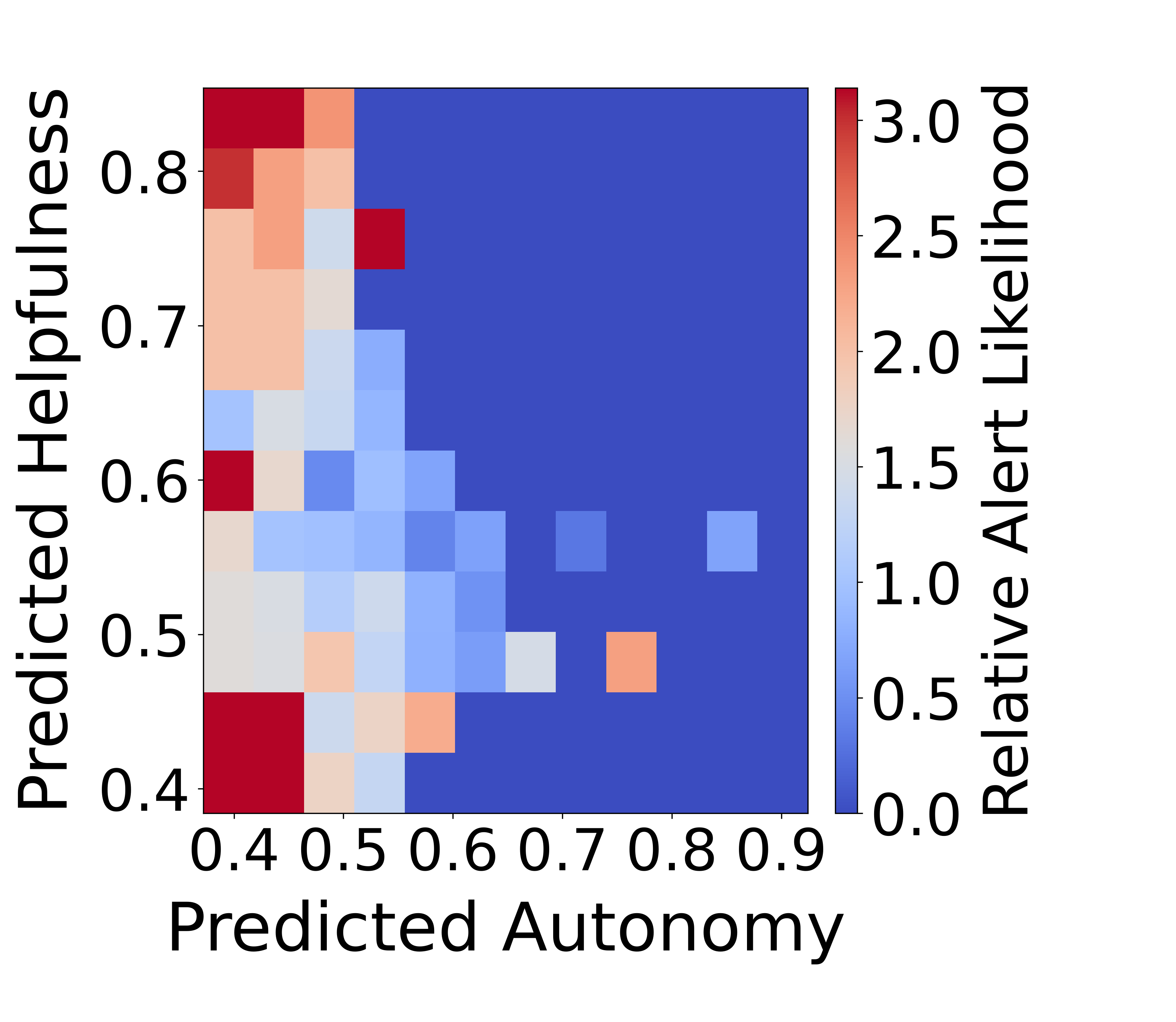}
        \caption{Relative alert frequency.}
        \label{fig:alert_heatmap}
    \end{subfigure}\hfill%
    \begin{subfigure}[t]{0.495\columnwidth}
        \centering
        \includegraphics[
            width=\linewidth
        ]{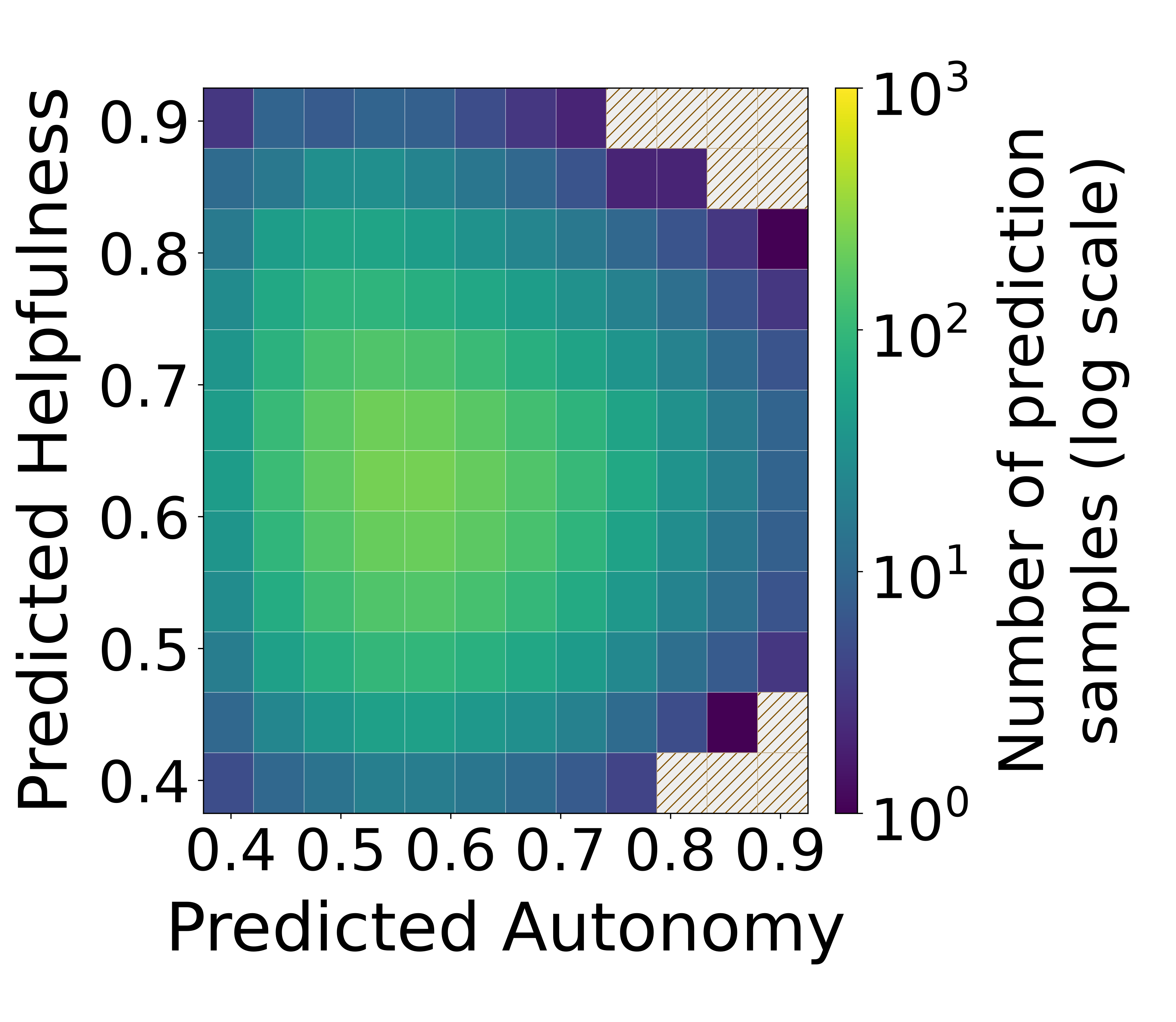}
        \caption{Sample occupancy.}
        \label{fig:occupancy}
    \end{subfigure}

    \vspace{-1mm}
    \caption{
    {
    Post-hoc characterization of the learned alert policy
    over predicted autonomy and helpfulness.
    (a) Relative alert frequency normalized by the
    overall alert rate; values above one indicate
    higher-than-average alert frequency.
    (b) Sample occupancy, showing the number of timestep-level predictions per cell (log scale). Hatched cells indicate zero occupancy and are interpolated in (a) for visualization only.
    }}
    \label{fig:visualization}
    \vspace{-2mm}
\end{figure}

Fig.~\ref{fig:visualization} { presents a post-hoc analysis of the Q-head's alert decisions in relation to the predicted autonomy and helpfulness scores.} Alerts are concentrated in regions of low predicted autonomy (left-side), indicating that the policy requests assistance when autonomous recovery is unlikely, while avoiding unnecessary alerts in high-autonomy states (right-side). Within low-autonomy regions, alert likelihood increases with predicted helpfulness (top-left), suggesting that the policy prioritizes situations where intervention is expected to be beneficial.

\section{Discussion}

This work demonstrates that robots can effectively learn when to request human assistance during exploration. In simulation, the proposed policy outperformed the threshold-based, score-based, and LinUCB baselines, while the physical user study showed improved exploration performance and reduced supervisory workload.

A key factor underlying this behavior is the model’s ability to estimate autonomy and intervention helpfulness. {These objectives provide additional supervision to the shared learned representation.} Their consistency across participants and alignment with robot behavior suggest that these representations generalize beyond the training setting and remain effective under real-world variability. The ablation results further show that the autonomy and helpfulness objectives provide complementary, condition-dependent benefits, with their combination yielding the most consistent overall performance. Moreover, the improvement over the score-based and LinUCB baselines suggests that the gains are not explained by the auxiliary predictions or context-dependent decision making alone, supporting the benefit of integrating these signals within a sequential RL policy.

The {post-hoc policy visualization
further characterizes} this interpretation. The heatmap analysis reveals that alert decisions are {associated with predicted
autonomy and helpfulness scores}. Alerts are predominantly {observed} in regions of low predicted autonomy, {corresponding to states where
autonomous progress is predicted to be limited}. Within these regions, alert likelihood increases with predicted helpfulness, suggesting that {alerts are more frequent in states
where human intervention is predicted to be beneficial}.

Unlike fixed heuristics (e.g., velocity thresholds), our adaptive approach reduces unnecessary requests, allowing operators to allocate attention effectively. Subjective results confirm a significant reduction in perceived workload in the learned alert condition. Interestingly, trust remained comparable across strategies. The slightly higher trust in threshold-based alerts likely stems from the consistent triggers that align with intuitive failure cues (e.g., stopping) that match users’ existing mental models of robot behavior~\cite{lee2004trust}.

Several limitations should be noted. First, the simulation assumes immediate responses to alerts, whereas real users may delay or ignore alerts, introducing variability in outcomes. Second, the physical study primarily involved a relatively small and primarily novice participant sample; future work should examine how experienced operators interact with adaptive alerting policies. Third, although the policy generalizes from simulation to real-world settings, evaluation was limited to a specific exploration task and environment. {
Finally, the physical study focused on mission-level outcomes, as the logs do not support reliable failure-type-specific alert and recovery comparisons.}

\section{Conclusions and Future Work}

This paper investigated how robots can learn when to request human assistance during exploration tasks. We proposed an RL-based alert policy that adapts help requests based on autonomous progress and past intervention outcomes. Across both simulation and real-world user studies, the learned policy improved exploration performance while issuing fewer alerts and reducing operator workload. Overall, this work highlights the importance of adaptive help-seeking for balancing autonomy and human supervision. Future work will
explore user-aware models, online adaptation, additional
intervention modalities, and multi-robot settings in which several robots
compete for limited operator attention.

\bibliographystyle{IEEEtran}

\bibliography{reference}

\end{document}